\pdfoutput=1%

\documentclass[11pt,a4paper]{article}

\usepackage[dvipsnames]{xcolor} %

\usepackage[bookmarks]{hyperref}
\usepackage[hyperref]{emnlp2018}
\usepackage{times}
\usepackage{latexsym}

\aclfinalcopy %

\usepackage{xparse}
\usepackage{amsmath,amssymb,mathtools}
\usepackage{booktabs,multirow}
\usepackage{amsbsy}

\usepackage{zi4}

\newcommand{\mathboldface}[1]{\mbox{\boldmath $#1$}}
\newcommand{\VEC}[1]{\mathboldface{#1}}

\NewDocumentCommand\Prob{m o}{%
{%
\mathbf{P}_{\!{#1}}^{}
\IfValueT{#2}{(#2)}
}
}

\usepackage{tikz}
\newcommand{\empheqpos}[2]{%
  \tikz[baseline]
  \node[fill=LimeGreen!45, rounded corners=3pt, anchor=base]
  (#1)
  {\ensuremath{#2}};}
\newcommand{\empheqneg}[2]{%
  \tikz[baseline]
  \node[fill=lightgray!45, rounded corners=3pt, anchor=base]
  (#1)
  {\ensuremath{#2}};}

\usepackage{microtype}
\AtBeginDocument{
\setlength\abovedisplayskip{7pt plus 2pt minus 4pt}
\setlength\belowdisplayskip{7pt plus 2pt minus 4pt}
\setlength\abovedisplayshortskip{\abovedisplayskip}
\setlength\belowdisplayshortskip{\belowdisplayskip}
}
\allowdisplaybreaks
\renewenvironment{quote}
{\list{}{\rightmargin=1em \leftmargin=1.8em}%
\item\relax}
{\endlist}
\usepackage{enumitem}
\setlist{noitemsep, topsep=1pt, leftmargin=*}

\usepackage[hang]{footmisc}

\title{Pointwise HSIC: A Linear-Time Kernelized Co-occurrence Norm\\for Sparse Linguistic Expressions}

\author{%
Sho Yokoi${}^{\,1, 2}$
\quad\hspace{-5pt}
Sosuke Kobayashi${}^{\,3}$
\quad\hspace{-5pt}
Kenji Fukumizu${}^{\,4}$
\quad\hspace{-5pt}
Jun Suzuki${}^{\,1, 2}$
\quad\hspace{-5pt}
Kentaro Inui${}^{\,1, 2}$
\\[5pt]
${}^{1}$ Tohoku University \quad
${}^{2}$ RIKEN Center for Advanced Intelligence Project \\
{\tt \{yokoi,\hspace{0em}jun.suzuki,\hspace{0em}inui\}@ecei.tohoku.ac.jp}
\\[5pt]
${}^{3}$ Preferred Networks, Inc. \quad
${}^{4}$ The Institute of Statistical Mathematics \\
{\tt sosk@preferred.jp}
\qquad\qquad\qquad
\hspace{-0.2em}{\tt fukumizu@ism.ac.jp}\hspace{0.4em}\qquad\quad
}

\date{}

\begin{document}
\maketitle
\begin{abstract}
In this paper, we propose a new kernel-based co-occurrence measure that can be applied to sparse linguistic expressions (e.g., sentences) with a very short learning time,
as an alternative to pointwise mutual information (PMI).
As well as deriving PMI from mutual information,
we derive this new measure from the Hilbert--Schmidt independence criterion (HSIC); thus, we call the new measure the pointwise HSIC (PHSIC).
PHSIC can be interpreted as a smoothed variant of PMI that allows various similarity metrics (e.g., sentence embeddings) to be plugged in as kernels.
Moreover, PHSIC can be estimated by simple and fast (linear in the size of the data) matrix calculations
regardless of whether we use linear or nonlinear kernels.
Empirically, in a dialogue response selection task, PHSIC is learned thousands of times faster than an RNN-based PMI while outperforming PMI in accuracy.
In addition, we also demonstrate that PHSIC is beneficial as a criterion of a data selection task for machine translation  owing to its ability to give high (low) scores to a consistent (inconsistent) pair with other pairs.
\end{abstract}

\section{Introduction}
\label{sec:introduction}
Computing the co-occurrence strength between two linguistic expressions is a fundamental task in natural language processing~(NLP).
For example,
in collocation extraction~\cite{ManningSchutze1999},
word bigrams are collected from corpora and then strongly co-occurring bigrams~(e.g., ``New York'') are found.
In dialogue response selection~\cite{Lowe2015},
pairs comprising a context and its response sentence are collected from dialogue corpora and the goal is to rank the candidate responses for each given context sentence.
In either case, a set of linguistic expression pairs $\mathcal D = \{(x_i,y_i)\}_{i=1}^n$ is first collected and then the co-occurrence strength of a (new) pair $(x,y)$ is computed.

Pointwise mutual information (PMI) \cite{Church1989} is frequently used to model the co-occurrence strength of linguistic expression pairs.
There are two typical types of PMI estimation (computation) method.
\begin{table}[tb]
\begin{center}
\footnotesize
\setlength{\tabcolsep}{1pt}
\begin{tabular}{l @{\hspace{2pt}} l c @{\hspace{2pt}} c}
\toprule
&
& \multicolumn{1}{r}{\hspace{-25pt}Robustness}\hspace{16pt}
& \multicolumn{1}{r}{\hspace{-15pt}Learning}
\\
&
& \multicolumn{1}{r}{\hspace{-25pt}to Sparsity}\hspace{16pt}
& \multicolumn{1}{r}{\hspace{-15pt}Time}
\\
\midrule
\textbf{PMI} & & & \\
$\empheqneg{}{\displaystyle
\!\log \frac{n \cdot c(x,y)}{\sum_{y'}\!c(x,y')\sum_{x'}\!c(x'\!,y)}
\!}$\vspace*{2pt} & Eq. \ref{eq:pmi_mle} & & $\pmb{\checkmark}$\hspace{5pt} \\
$\empheqneg{}{\displaystyle\!
\log \frac{\widehat{\mathbf{P}}_{\mathrm{RNN}}(y|x)}{\widehat{\mathbf{P}}_{\mathrm{RNN}}(y)}
\!}$\vspace*{0.5pt} & Eq. \ref{eq:pmi_rnn} & \hspace{4pt}$\pmb{\checkmark}$\hspace{10pt} & \\
\midrule
\textbf{PHSIC} & & \\
$\empheqpos{}{\!
(\phi(x) \!-\! \overline{\phi(x)})^{\!\top}\widehat{C}_{\!XY}^{}(\psi(y) \!-\! \overline{\psi(y)})
\!}$\vspace*{2pt} & Sec. \ref{sec:estimation_RKHS} & \hspace{4pt}$\pmb{\checkmark}$\hspace{10pt} & $\pmb{\checkmark}$\hspace{5pt} \\
$\empheqpos{}{\!
(\VEC{a} \!-\! \overline{\VEC{a}})^\top
\widehat C_{\mathrm{ICD}}^{}
(\VEC{b} \!-\! \overline{\VEC{b}})
\!}$\vspace*{0.5pt} & Sec. \ref{sec:estimation_data_space} & \hspace{4pt}$\pmb{\checkmark}$\hspace{10pt} & $\pmb{\checkmark}$\hspace{5pt} \\
\bottomrule
\end{tabular}
\end{center}

\caption{The proposed co-occurrence norm, PHSIC, eliminates the trade-off between robustness to data sparsity and learning time,
which PMI has~(Section~\ref{sec:introduction}).}
\label{table:contribution}
\end{table}
One is a counting-based estimator using maximum likelihood estimation, sometimes with smoothing techniques, for example,
\begin{align}
\widehat{\mathrm{PMI}}_{\mathrm{MLE}}(x,y;\mathcal D)
&\!=\!
\empheqneg{}{\displaystyle
\!\log \frac{n \cdot c(x,y)}
{\sum_{y'}\!c(x,y')\sum_{x'}\!c(x'\!,y)}
\!}
\,\text{,}\!\label{eq:pmi_mle}
\end{align}
where $c(x,y)$ denotes the frequency of the pair~$(x,y)$ in given data $\mathcal D$.
This is easy to compute and is commonly used to measure co-occurrence between words, such as in collocation extraction\footnote{%
In collocation extraction, simple counting $c(x,y) \propto \widehat{\mathbf P}(x,y)$, rather than PMI, ranks undesirable function-word pairs (e.g., ``of the'') higher~\cite{ManningSchutze1999}.%
};
however, when data $\mathcal D$ is sparse, i.e., when $x$ or $y$ is a phrase or sentence, this approach is unrealistic.
The second method uses recurrent neural networks~(RNNs).
\citet{Li2016} proposed to employ PMI to suppress \emph{dull} responses for utterance generation in dialogue systems%
\footnote{In dialogue response selection or generation, a simple conditional probability $\widehat{\mathbf P}(y|x)$, rather than PMI, ranks \emph{dull} responses (e.g., ``I don't know.'') higher~\cite{Li2016}.}.
They estimated $\mathbf{P}(y)$ and $\mathbf{P}(y|x)$ using RNN language models and estimated PMI as follows:
\begin{align}
\widehat{\mathrm{PMI}}_{\mathrm{RNN}}^{}(x,y;\mathcal D) =
\empheqneg{}{\displaystyle\!
\log \frac{\widehat{\mathbf{P}}_{\mathrm{RNN}}(y|x)}{\widehat{\mathbf{P}}_{\mathrm{RNN}}(y)}
\!}
\,\text{.}
\label{eq:pmi_rnn}
\end{align}
This way of estimating PMI is applicable to sparse language expressions;
however, learning RNN language models is 
computationally costly.

To eliminate this trade-off between robustness to data sparsity and learning time, in this study we propose a new kernel-based co-occurrence measure, which we call the \emph{pointwise Hilbert--Schmidt independence criterion (PHSIC)} (see Table~\ref{table:contribution}).
Our contributions are as follows:
\begin{itemize}
\item We formalize PHSIC, which is derived from HSIC~\cite{Gretton2005}, a kernel-based dependence measure, in the same way that PMI is derived from mutual information (Section~\ref{sec:definition_phsic}).
\item We give an intuitive explanation why PHSIC is robust to data sparsity. PHSIC is a ``smoothed variant of PMI'', which allows various similarity metrics to be plugged in as kernels (Section~\ref{sec:phsic_as_smoothed_pmi}).
\item We propose fast estimators of PHSIC, which are reduced to a simple and fast matrix calculation
regardless of whether we use linear or nonlinear kernels
(Section~\ref{sec:estimator_of_phsic}).
\item We empirically confirmed the effectiveness of PHSIC, i.e., its robustness to data sparsity and learning time, in two different types of experiment, a dialogue response selection task and a data selection task for machine translation (Section~\ref{sec:experiments}).
\end{itemize}

\section{Problem Setting}
Let $X$ and $Y$ denote random variables on $\mathcal X$ and $\mathcal Y$, respectively.
In this paper,
we deal with the tasks of
taking a set of linguistic expression pairs
\begin{align}
\mathcal D = \{(x_i,y_i)\}_{i=1}^n \underset{\text{i.i.d.}}{\sim} \Prob{XY}
\text{,}
\end{align}
which is regarded as a set of i.i.d.\ samples drawn from a joint distribution $\Prob{XY}$,
and then measuring the ``co-occurrence strength'' for each given pair $(x,y) \in \mathcal X\times\mathcal Y$.
Such tasks include collocation extraction and dialogue response selection~(Section~\ref{sec:introduction}).

\section{Pointwise HSIC}
\label{sec:definition_phsic}
In this section,
we give the formal definition of PHSIC,
a new kernel-based co-occurrence measure.
We show a summary of this section in Table~\ref{table:mi_pmi_hsic_phsic}.
Intuitively, PHSIC is a ``kernelized variant of PMI.''
\begin{table*}[tb]
\setlength{\tabcolsep}{0.9em}
\begin{center}
\footnotesize
\begin{tabular}{c l l}
\toprule
&
\multicolumn{1}{l}{
\textbf{Dependence Measure}
}
& \multicolumn{1}{l}{
\textbf{Co-occurrence Measure}
}\\[2pt]
& \quad
the dependence between $X$ and $Y$
& \quad
the contribution of $(x,y)$
\\
& \quad
(the difference between $\Prob{XY}$ and $\Prob{X}\Prob{Y}$)
& \quad
to the dependence between $X$ and $Y$
\\
\midrule
\textbf{MI \& PMI} &
$\begin{aligned}
&\mathrm{MI}(X,Y) = \mathrm{KL}[\Prob{XY} \| \Prob{X}\Prob{Y}]\\
&= \underset{(x,y)}{\mathbf E}\Biggl[
\empheqneg{}{\displaystyle
\log\frac{\Prob{XY}[x,y]}{\Prob{X}[x]\Prob{Y}[y]}
}
\Biggr]
\end{aligned}$
&
$\begin{aligned}
&\mathrm{PMI}(x,y;X,Y)\\
&= \empheqneg{}{\displaystyle
\log\frac{\Prob{XY}[x,y]}{\Prob{X}[x]\Prob{Y}[y]}
}
\end{aligned}$
\\
\midrule
\textbf{HSIC \& PHSIC} &
$\begin{aligned}
&\mathrm{HSIC}(X,Y;k,\ell)
= \mathrm{MMD}_{k,\ell}^2[\Prob{XY}, \Prob{X}\Prob{Y}]\\[-1.5pt]
&= \underset{(x,y)}{\mathbf E}\Bigl[
\empheqpos{}{\displaystyle
(\phi(x) - m_X^{})^\top C_{XY}^{} (\psi(y) - m_Y^{})
}
\Bigr]\\[-3pt]
&= \underset{(x,y)}{\mathbf E}\Bigl[
\empheqpos{}{\displaystyle
\underset{(x',y')}{\mathbf E}[
\widetilde k(x,x')\widetilde \ell(y,y')
]
}
\Bigr]
\end{aligned}$
&
$\begin{aligned}
&\mathrm{PHSIC}(x,y;X,Y,k,\ell)\\
&= \empheqpos{}{\displaystyle
(\phi(x) - m_X^{})^\top C_{XY}^{} (\psi(y) - m_Y^{})
}
\\
&= \empheqpos{}{\displaystyle
\underset{(x',y')}{\mathbf E}[
\widetilde k(x,x')\widetilde \ell(y,y')
]
}
\end{aligned}$
\\
\bottomrule
\end{tabular}
\end{center}

\caption{Relationship between the mutual information~(MI), the pointwise mutual information~(PMI), the Hilbert--Schmidt independence criterion~(HSIC), and the pointwise HSIC~(PHSIC).
As well as defining PMI as the contribution to MI, we define PHSIC as the contribution to HSIC.
In short, PHSIC is a ``kernelized PMI'' (Section~\ref{sec:definition_phsic}).}
\label{table:mi_pmi_hsic_phsic}
\end{table*}

\subsection{Dependence Measure}
\label{sec:dependence_measure}
As a preliminary step,
we introduce the simple concept of dependence (see \textbf{Dependence Measure} in Table~\ref{table:mi_pmi_hsic_phsic}).
Recall that random variables $X$ and $Y$ are \emph{independent} if and only if the joint probability density $\Prob{XY}$ and the product of the marginals $\Prob{X}\Prob{Y}$ are equivalent.
Therefore, we can measure the \emph{dependence} between random variables $X$ and $Y$
via the difference between $\Prob{XY}$ and $\Prob{X}\Prob{Y}$.

Both the mutual information and the Hilbert--Schmidt independence criterion, to be described below, are such dependence measures.

\subsection{MI and PMI}
\label{sec:mi_and_pmi}
We briefly review the well-known mutual information and PMI (see \textbf{MI \& PMI} in Table~\ref{table:mi_pmi_hsic_phsic}).

The \emph{mutual information~(MI)}\footnote{Conventionally, mutual information is denoted by $I(X;Y)$;
in this paper, however, for notational consistency, mutual information is denoted by $\mathrm{MI}(X,Y)$.} between two random variables $X$ and $Y$ is defined by
\begin{align}
\mathrm{MI}(X,Y) := \mathrm{KL}[\Prob{XY} \| \Prob{X}\Prob{Y}]
\label{eq:mi_kl}
\end{align}
\cite{Cover2006}, where $\mathrm{KL}[\cdot\|\cdot]$ denotes the Kullback--Leibler (KL) divergence.
Thus, $\mathrm{MI}(X,Y)$ is the degree of dependence
between $X$ and $Y$
measured by the KL divergence
between $\Prob{XY}$ and $\Prob{X}\Prob{Y}$.

Here,
by definition of the KL divergence,
MI can be represented in the form of the expectation over $\Prob{XY}$,
i.e., the summation over all possible pairs $(x,y)\in \mathcal X\!\times\!\mathcal Y$:
\begin{align}
\mathrm{MI}(X,Y)
&= \underset{(x,y)}{\mathbf E}\Biggl[
\empheqneg{}{\displaystyle
\log\frac{\Prob{XY}[x,y]}{\Prob{X}[x]\Prob{Y}[y]}
}
\Biggr]\label{eq:mi_as_expectation_of_pmi}
\text{.}
\end{align}
The shaded part in Equation~\eqref{eq:mi_as_expectation_of_pmi} is actually the \emph{pointwise mutual information (PMI)} \cite{Church1989}:
\begin{align}
\mathrm{PMI}(x,y;X,Y) := \empheqneg{}{\displaystyle
\log\frac{\Prob{XY}[x,y]}{\Prob{X}[x]\Prob{Y}[y]}
}\label{eq:pmi_as_contribution_of_mi}
\,\text{.}
\end{align}
Therefore, $\mathrm{PMI}(x,y)$ can be thought of as the contribution of $(x,y)$ to $\mathrm{MI}(X,Y)$.

\subsection{HSIC and PHSIC}
\label{sec:hsic_and_phsic}
As seen in the previous section, PMI can be derived from MI.
Here,
we consider replacing MI with
the Hilbert--Schmidt independence criterion (HSIC).
Then, in analogy with the relationship between PMI and MI, we derive PHSIC from HSIC
(see \textbf{HSIC \& PHSIC} in Table~\ref{table:mi_pmi_hsic_phsic}).

Let $k\colon\mathcal X\times\mathcal X\to \mathbb R$ and $\ell\colon\mathcal Y\times\mathcal Y\to \mathbb R$ denote positive definite kernels on $\mathcal X$ and $\mathcal Y$, respectively~(intuitively, they are similarity functions between linguistic expressions).
The \emph{Hilbert--Schmidt independence criterion (HSIC)} \cite{Gretton2005}, a kernel-based dependence measure,
is defined by
\begin{align}
\mathrm{HSIC}(X,Y\!;k,\ell)
& \textstyle \!:=\! \mathrm{MMD}_{k,\ell}^2[\Prob{XY}
\vspace{-0.5pt}, \Prob{X}\Prob{Y}]_{}\label{eq:hsic_definition_mmd}
\text{,}\!
\end{align}
where $\mathrm{MMD}[\cdot, \cdot]$ denotes the maximum mean discrepancy (MMD)~\cite{Gretton2012},
which measures the difference between random variables on a kernel-induced feature space.
Thus,
$\mathrm{HSIC}(X,Y;k,\ell)$ is the degree of dependence
between $X$ and $Y$
measured by the MMD between $\Prob{XY}$ and $\Prob{X}\Prob{Y}$,
while MI is
measured by the KL divergence
(Equation~\eqref{eq:mi_kl}).

Analogous to MI in Equation~\eqref{eq:mi_as_expectation_of_pmi},
HSIC can be represented in the form of the expectation on $\Prob{XY}$
by a simple deformation:
\begin{align}
&\mathrm{HSIC}(X,Y;k,\ell)\nonumber\\
&= \underset{\mathclap{(x,y)}}{\mathbf E}\hspace{4pt}\Bigl[
\empheqpos{}{\displaystyle
\!(\phi(x) \!-\! m_X^{})^{\!\top\!} C_{XY}^{} (\psi(y) \!-\! m_Y^{})
}
\Bigr]\label{eq:hsic_population_RKHS}
\\[-4.5pt]
&= \underset{\mathclap{(x,y)}}{\mathbf E}
\hspace{4pt}\Bigl[
\empheqpos{}{\displaystyle
\underset{(x',y')}{\mathbf E}[
\widetilde k(x,x')\widetilde \ell(y,y')
]
}
\Bigr]\label{eq:hsic_population_kernel}
\text{,}
\end{align}
where
\begin{align}
& \phi(x) := k(x,\cdot)
\text{,}
\quad \psi(y) := \ell(y,\cdot)
\text{,}
\\
&
m_X^{} := \mathbf{E}_x[\phi(x)]
\text{,}
\quad
m_Y^{} := \mathbf{E}_y[\psi(y)]
\text{,}
\\
& C_{XY}^{} := \underset{\mathclap{(x,y)}}{\mathbf E}\hspace{4pt}\Bigl[
(\phi(x) -
m_X^{}
)
(\psi(y) -
m_Y^{}
)^{\!\top}\!
\Bigr]
\text{,}\!
\\[-2pt]
& \widetilde k(x,x')
:= k(x,x') - \mathbf E_{x'}[k(x,x')]\nonumber\\
&\qquad\quad\, - \mathbf E_{x}[k(x,x')] + \mathbf E_{x,x'}[k(x,x')]
\text{.}\!\label{eq:centered_kernel}
\end{align}
At first glance, these equations are somewhat complicated; however, the estimators of PHSIC we actually use are reduced to a simple matrix calculation in Section \ref{sec:estimator_of_phsic}.
Unlike MI in Equation~\eqref{eq:mi_as_expectation_of_pmi}, HSIC has two representations:
Equation~\eqref{eq:hsic_population_RKHS} is the representation in feature space
and
Equation~\eqref{eq:hsic_population_kernel} is the representation in data space.

Similar to the relationship between MI and PMI (Section \ref{sec:mi_and_pmi}),
we define the \emph{pointwise Hilbert--Schmidt independence criterion~(PHSIC)} by the shaded parts in Equations~\eqref{eq:hsic_population_RKHS} and \eqref{eq:hsic_population_kernel}:
\begin{align}
&\mathrm{PHSIC}(x,y;X,Y,k,\ell)\nonumber\\
&:=
\empheqpos{}{\displaystyle
(\phi(x) - m_X^{})^\top C_{XY} (\psi(y) - m_Y^{})
}\label{eq:phsic_population_RKHS}
\\
&\hspace{3pt}=
\empheqpos{}{\displaystyle
\underset{(x',y')}{\mathbf E}[\widetilde k(x,x')\widetilde \ell(y,y')]
}\label{eq:phsic_population_kernel}
\,\text{.}\!
\end{align}
Namely, $\mathrm{PHSIC}(x,y)$ is defined as the contribution of $(x,y)$ to $\mathrm{HSIC}(X,Y)$.

In summary,
we define PHSIC such that ``MI:PMI = HSIC:PHSIC'' holds (see Table~\ref{table:mi_pmi_hsic_phsic}).

\begin{table*}[tb]
\newcommand{\veq}{\mathrel{\rotatebox{90}{$\strut =$}}}
\newcommand{\vneq}{\mathrel{\rotatebox{90}{$\strut \ne$}}}
\newcommand{\vapprox}{\mathrel{\rotatebox{90}{$\strut \approx$}}}
\newcommand{\vnapprox}{\mathrel{\rotatebox{90}{$\strut \not\approx$}}}
\newcommand{\EMPHX}[1]{%
  \tikz[baseline]
  \node[anchor=base, fill=WildStrawberry!50, circle, inner sep=0pt, minimum size=1.3em, style={font=\vphantom{Ag}}]
  {\ensuremath{#1}};}
\newcommand{\EMPHY}[1]{%
  \tikz[baseline]
  \node[anchor=base, fill=YellowOrange!50, circle, inner sep=0pt, minimum size=1.3em, style={font=\vphantom{Ag}}]
  {\ensuremath{#1}};}
\newcommand{\EMPHXAPPROX}[1]{%
  \tikz[baseline]
  \node[anchor=base, fill=WildStrawberry!20, circle, inner sep=0pt, minimum size=1.3em, style={font=\vphantom{Ag}}]
  {\ensuremath{#1}};}
\newcommand{\EMPHYAPPROX}[1]{%
  \tikz[baseline]
  \node[anchor=base, fill=YellowOrange!20, circle, inner sep=0pt, minimum size=1.3em, style={font=\vphantom{Ag}}]
  {\ensuremath{#1}};}

\begin{center}
\small
  \setlength{\tabcolsep}{0.5pt}
  \begin{tabular}{l | c c}
    \toprule
    & add scores
    & deduct scores
    \\\midrule
    {
    $\displaystyle\widehat{\mathrm{PMI}}(x,y;\mathcal D) =
\empheqneg{}{\displaystyle\!
\log \frac{n\cdot \sum_i \mathbb I[x \!=\! x_i \land y \!=\! y_i]}{\sum_i \mathbb I[x \!=\! x_i] \sum_i \mathbb I[y \!=\! y_i]}
\!}$
}
    & $\begin{array}{r @{} c @{} c @{} c @{} l}
         (
         &\EMPHX{x}
         &,
         &\EMPHY{y}
         &)\\
         &\veq & &\veq &\\[-1.3ex]
         \mathcal D = \{\dots,(
         &\EMPHX{x_i}
         &,
         &\EMPHY{y_i}
         &),\dots\}
       \end{array}$
    & $\begin{array}{r @{} c @{} c @{} c @{} c @{} c @{} c @{} c @{} c @{} c @{} l}
         (
         &\EMPHX{x}
         &,
         &\EMPHY{y}
         &) && (
         &\EMPHX{x}
         &,
         &\EMPHY{y}
         &)\\
         &\veq & &\vneq &&&&\vneq & &\veq &\\[-1.3ex]
         \{\dots,(
         &\EMPHX{x_i}
         &,
         &y_i
         &) &,\dots, &(
         &x_i
         &,
         &\EMPHY{y_i}
         &),\dots\}\!
      \end{array}$
    \\\midrule
    \,$\displaystyle\widehat{\mathrm{PHSIC}}(x,y;\mathcal D,k,\ell) =
\empheqpos{}{\!\textstyle
\frac 1 n \!\sum_i \widehat{\widetilde k}(x,x_i)\widehat{\widetilde\ell}(y,y_i)\!
}$\,\,
    & $\begin{array}{r @{} c @{} c @{} c @{} c @{} c @{} c @{} c @{} c @{} c @{} l}
         (
         &\EMPHX{x}
         &,
         &\EMPHY{y}
         &) && (
         &\EMPHX{x}
         &,
         &\EMPHY{y}
         &)\\
         &\vapprox & &\vapprox &&&&\vnapprox & &\vnapprox &\\[-1.3ex]
         \{\dots,(
         &\EMPHXAPPROX{x_i}
         &,
         &\EMPHYAPPROX{y_i}
         &) &,\dots, &(
         &x_i
         &,
         &y_i
         &),\dots\}
      \end{array}$
    & $\begin{array}{r @{} c @{} c @{} c @{} c @{} c @{} c @{} c @{} c @{} c @{} l}
         (
         &\EMPHX{x}
         &,
         &\EMPHY{y}
         &) && (
         &\EMPHX{x}
         &,
         &\EMPHY{y}
         &)\\
         &\vapprox & &\vnapprox &&&&\vnapprox & &\vapprox &\\[-1.3ex]
         \{\dots,(
         &\EMPHXAPPROX{x_i}
         &,
         &y_i
         &) &,\dots, &(
         &x_i
         &,
         &\EMPHYAPPROX{y_i}
         &),\dots\}\!
      \end{array}$
    \\\bottomrule
    \end{tabular}
  \end{center}

\caption{Comparison of estimators of PMI and PHSIC in terms of methods of matching the given $(x,y)$ and the observed $(x_i,y_i)$ in $\mathcal D$.
PMI matches them in an exact manner, while PHSIC smooths the matching using kernels.
Therefore, PHSIC is expected to be robust to data sparsity~(Section~\ref{sec:phsic_as_smoothed_pmi}).%
}
\label{table:phsic_as_smoothed_pmi}
\end{table*}
\section{PHSIC as Smoothed PMI}
\label{sec:phsic_as_smoothed_pmi}
This section gives an intuitive explanation for the first feature of PHSIC, i.e., the robustness to data sparsity, using Table~\ref{table:phsic_as_smoothed_pmi}.
In short, we show that PHSIC is a ``smoothed variant of PMI.''

First,
the maximum likelihood estimator of PMI in Equation~\eqref{eq:pmi_mle}
can be rewritten as
\begin{align}
\!\!\!\!\widehat{\mathrm{PMI}}(x,y;\!\mathcal D)
\!=\!
\empheqneg{}{\displaystyle\!
\log\!\frac{n\cdot \sum_i \!\mathbb I[x\!=\!x_i \land y\!=\!y_i]}{\sum_i \!\mathbb I[x\!=\!x_i] \sum_i \!\mathbb I[y\!=\!y_i]}%
\!}
\,\text{,}\!\!\label{eq:pmi_mle_compare}
\end{align}
where $\mathbb I[\text{condition}] = 1$ if the condition is true and $\mathbb I[\text{condition}] = 0$ otherwise.
According to Equation~\eqref{eq:pmi_mle_compare},
$\widehat{\mathrm{PMI}}(x,y)$ is the amount computed by repeating the following operation~(see the first row in Table~\ref{table:phsic_as_smoothed_pmi}):
\begin{quote}\it
collate the given $(x,y)$ and the observed $(x_i,y_i)$ in $\mathcal D$ in order,
and add the scores if $(x,y)$ and $(x_i,y_i)$ match exactly
or deduct the scores
if either the $x$ side or the $y$ side (but nor both) matches.
\end{quote}

Moreover,
an estimator of PHSIC
in data space
(Equation \eqref{eq:phsic_population_kernel})
is
\begin{align}
\!\widehat{\mathrm{PHSIC}}(x,y;\mathcal D,k,\ell) \!=\!
\empheqpos{}{\!
\frac 1 n \!\sum_i \widehat{\widetilde k}(x,x_i)\widehat{\widetilde\ell}(y,y_i)%
\!}
\,\text{,}\!\!\label{eq:phsic_mle_compare}
\end{align}
where
$\widehat{\widetilde k}(\cdot,\cdot)$ and
$\widehat{\widetilde \ell}(\cdot,\cdot)$
are similarity functions centered on the data\footnote{%
To be exact, $\widehat{\widetilde k}(x,x') :=
k(x,x')
- \frac 1 n \!\sum_{j=1}^n k(x,x_j)%
- \frac 1 n \!\sum_{i=1}^n k(x_i,x') 
+ \frac 1 {n^2} \!\sum_{i=1}^n\!\sum_{j=1}^n k(x_i,x_j)$,
which is an estimator of the centered kernel $\widetilde k(x,x')$ in Equation~\eqref{eq:centered_kernel}.
}.
According to Equation~\eqref{eq:phsic_mle_compare},
$\widehat{\mathrm{PHSIC}}(x,y)$ is the amount computed by repeating the following operation~(see the second row in Table~\ref{table:phsic_as_smoothed_pmi}):
\begin{quote}\it
collate the given $(x,y)$ and the observed $(x_i,y_i)$ in $\mathcal D$ in order,
and add the scores if the similarities on the $x$ and $y$ sides are both higher
(both $\widehat{\widetilde k}(x,x_i)>0$ and $\widehat{\widetilde \ell}(y,y_i)>0$ hold)\footnote{%
In addition, the scores are added if the similarity on the $x$ side and that on the $y$ side are both lower, that is, if $\widehat{\widetilde k}(x,x_i)<0$ and $\widehat{\widetilde \ell}(y,y_i)<0$ hold.%
}
or deduct the scores if the similarities on either the $x$ or $y$ sides are similar but those on the other side are not similar.
\end{quote}

As described above,
when comparing the estimators of PMI and PHSIC from the viewpoint of ``methods of matching the given $(x,y)$ and the observed $(x_i,y_i)$,''
it is understood that PMI matches them in an exact manner,
while PHSIC smooths the matching using kernels~(similarity functions).

With this mechanism, even for completely unknown pairs, it is possible to estimate the co-occurrence strength by referring to observed pairs through the kernels.
Therefore, PHSIC is expected to be robust to data sparsity
and can be applied to phrases and sentences.

\paragraph{Available Kernels for PHSIC}
\label{sec:kernel_available}
In NLP,
a variety of similarity functions (i.e., positive definite kernels) are available.
We can freely utilize such resources, such as cosine similarity between sentence embeddings.
For a more detailed discussion, see Appendix \ref{sec:available_kernels_appendix}.

\section{Empirical Estimators of PHSIC}
\label{sec:estimator_of_phsic}
Recall that we have two types of empirical estimator of PMI, the maximum likelihood estimator (Equation \eqref{eq:pmi_mle}) and the RNN-based estimator (Equation \eqref{eq:pmi_rnn}).
In this section, we describe how to rapidly estimate PHSIC from data.
When using the linear kernel or cosine similarity
(e.g., cosine similarity between sentence embeddings),
PHSIC
can be efficiently estimated in feature space~(Section~\ref{sec:estimation_RKHS}).
When using a nonlinear kernel such as the Gaussian kernel,
PHSIC
can also be estimated efficiently in data space via a simple matrix decomposition~(Section~\ref{sec:estimation_data_space}).

\subsection{Estimation Using Linear Kernel or Cosine\!}
\label{sec:estimation_RKHS}
When using the linear kernel or cosine similarity,
the estimator of PHSIC in feature space \eqref{eq:phsic_population_RKHS}
is as follows:
\newpage
\begin{align}
&\!\!\widehat{\mathrm{PHSIC}}_{\mathrm{feature}}(x,y;\mathcal D,k,\ell)\nonumber\\
&\!\!\hphantom{\widehat{C}_{XY}^{}} \!\,=\!
\empheqpos{}{\!
(\phi(x) \!-\! \overline{\phi(x)})^{\!\top}\widehat{C}_{\!XY}^{}(\psi(y) \!-\! \overline{\psi(y)})
\!}
\,\text{,}\!\label{eq:phsic_RKHS}
\end{align}
where
\begin{align}
&\phi(x) =
\!\begin{cases}
x                &\text{($k(x,x') = x^\top x'$)}\\
x/\lVert x\rVert &\text{($k(x,x') = \cos(x,x')$)}
\end{cases}\label{eq:feature_maps_linear_cosine}\!\text{,}\!
\\
&\overline{\phi(x)} \!:=\! \frac 1 n \!\sum_{i=1}^n \phi(x_i)
\text{,}
\quad\!\overline{\psi(y)} \!:=\! \frac 1 n \!\sum_{i=1}^n \psi(y_i)
\text{,}
\\
&\widehat{C}_{XY}^{}\!:=\! \frac 1 n \sum_{i=1}^n \phi(x_i)\psi(y_i)^{\!\top} \!- \overline{\phi(x)} \,\overline{\psi(y)}^{\top\!}
\text{.}
\end{align}
Generally in kernel methods,
a feature map $\phi(\cdot)$ induced by a kernel $k(\cdot,\cdot)$ is unknown or high-dimensional
and it is difficult to compute estimated values in feature space\footnote{%
One of the characteristics of kernel methods is
that an intractable estimation in feature space is replaced
with an efficient estimation in data space.%
}.
However,
when we use the linear kernel or cosine similarity, feature maps can be explicitly determined~(Equation \eqref{eq:feature_maps_linear_cosine}).

\paragraph{Computational Cost}
When learning Equation~\eqref{eq:phsic_RKHS} with feature maps $\phi\colon\mathcal X\to\mathbb R^d$ and $\psi\colon\mathcal Y\to\mathbb R^d$,
computing the vectors $\overline{\phi(x)}, \overline{\psi(y)}\in\mathbb R^d$ and the matrix $\widehat{C}_{XY}^{}\in\mathbb R^{d\times d}$ takes $\mathcal O(nd^2)$ time and $\mathcal O(nd)$ space (linear in the size of the input, $n$).
When estimating $\mathrm{PHSIC}(x,y)$,
computing $\phi(x),\psi(y)\in\mathbb R^d$ and Equation~\eqref{eq:phsic_RKHS} takes $\mathcal O(d^2)$ time (constant; does not depend on the size of the input, $n$).

\subsection{Estimation Using Nonlinear Kernels}
\label{sec:estimation_data_space}
When using a nonlinear kernel such as the Gaussian kernel,
it is necessary to estimate PHSIC in data space.
Using a simple matrix decomposition,
this can be achieved with the same computational cost as the estimation in feature space.
See Appendix \ref{sec:estimation_data_space_appendix} for a detailed derivation.

\section{Experiments}
\label{sec:experiments}
In this section, we provide empirical evidence for the greater effectiveness of PHSIC
than PMI,
i.e., a very short learning time and robustness to data sparsity.
Among the many potential applications of PHSIC, 
we choose two fundamental scenarios, (re-)ranking/classification and data selection.
\begin{itemize}
\item In the ranking/classification scenario (measuring the co-occurrence strength of \emph{new data pairs} with reference to observed pairs), PHSIC is applied as a criterion for the dialogue response selection task (Section~\ref{sec:experiment_dialogue}).
\item In the data selection/filtering scenario (ordering the entire set of \emph{observed data pairs} according to the co-occurrence strength), PHSIC
is also applied as a criterion for data selection
in the context of machine translation (Section~\ref{sec:experiment_data_selection}).
\end{itemize}

\subsection{PHSIC Settings}
To take advantage of recent developments in representation learning,
we used several pre-trained models for encoding sentences into vectors and several kernels between these vectors for PHSIC.
\label{sec:experiments_proposed}
\paragraph{Encoders}
As sentence encorders,
we used two pre-trained models without fine-tuning.
First, the sum of the word vectors
effectively represents a sentence~\cite{Mikolov2013}:
\begin{align}
\textstyle
\VEC{x} \!=\! \sum_{w\in x}\!\mathrm{vec}(w),
\quad
\VEC{y} \!=\! \sum_{w\in y}\!\mathrm{vec}(w)
\text{.}
\end{align}
For $\mathrm{vec}(\cdot)$,
we used the pre-trained \textbf{\texttt{fastText}} model\footnote{\url{https://fasttext.cc/docs/en/english-vectors.html}, \url{https://fasttext.cc/docs/en/crawl-vectors.html}},
which is a high-accuracy and popular word embedding model~\cite{Bojanowski2017};
models in 157 languages are publicly distributed~\cite{Grave2018}.
Second, we also used a DNN-based sentence encoder, called the universal sentence encoder \cite{Cer2018}, which utilizes the deep averaging network (\textbf{\texttt{DAN}}) \cite{Iyyer2015}.
The pre-trained model for English sentences we used is publicly available\footnote{\url{https://www.tensorflow.org/hub/modules/google/universal-sentence-encoder/1}}.

\paragraph{Kernels}
As kernels between these vectors, we used cosine similarity (\textbf{\texttt{cos}})
\begin{align}
k(\VEC{x},\VEC{x}') = \cos(\VEC{x},\VEC{x}')
\end{align}
and the Gaussian kernel (also known as the radial basis function kernel; \textbf{\texttt{RBF}} kernel)
\begin{align}
k(\VEC{x},\VEC{x}')
&= \exp\left(- \frac{\lVert \VEC{x} -\VEC{x}' \rVert_2^2}{2\sigma^2} \right)
\text{,}
\end{align}
and similarly for $\ell(\VEC{y},\VEC{y}')$.
The experiments are ran with hyperparameter $\sigma = 1.0$ for the RBF kernel, and $d = 100$ for incomplete Cholesky decomposition (for more detail, see Section \ref{sec:estimation_data_space_appendix}).

\subsection{Ranking: Dialogue Response Selection}
\label{sec:experiment_dialogue}
In the first experiment, we applied PHSIC as a ranking criterion of the task of dialogue response selection~\cite{Lowe2015};
in the task, pairs comprising a context (previous utterance sequence) and its response are collected from dialogue corpora and the goal is to rank the candidate responses for each given context sentence.

The task entails sentence sequences (very sparse linguistic expressions);
moreover, \citet{Li2016} pointed out that (RNN-based) PMI has a positive impact on suppressing \emph{dull} responses (e.g., ``I don't know.'') in dialogue systems.
Therefore, PHSIC, another co-occurrence measure, is also expected to be effective for this.
With this setting, where the validity of PMI is confirmed, we investigate whether PHSIC can replace RNN-based PMI in terms of both learning time and robustness to data sparsity.

\subsubsection*{Experimental Settings}
\paragraph{Dataset}
For the training data, we gathered approximately $5\times 10^5$ reply chains from Twitter, following \citet{Sordoni2015AResponses}\footnote{%
We collected tweets after 2017 for our training set to avoid duplication with the test set, which contains tweets from the year 2012.}.
In addition, we randomly selected $\{10^3, 10^4, 10^5\}$ reply chains from that dataset. Using these small subsets, we confirmed the effect of the difference in the size of the training set (data sparseness) on the learning time and predictive performance.

For validation and test data, we used a small (approximately $2000$ pairs each) but highly reliable dataset created by \citet{Sordoni2015AResponses}\footnote{\url{https://www.microsoft.com/en-us/download/details.aspx?id=52375}},
which consists only of conversations given high scores by human annotators.
Therefore, this set was not expected to include \textit{dull} responses.

For each dataset, we converted each context-message-response triple into a context-response pair by concatenating the context and message following \citet{Li2016}.
In addition, to convert the test set (positive examples) to ten-choice multiple-choice questions, we shuffled the combinations of context and response to generate pseudo-negative examples.

\paragraph{Evaluation Metrics}
We adopted the following evaluation metrics for the task:
(i) \texttt{ROC-AUC} (the area under the receiver operating characteristic curve),
(ii) \texttt{MRR} (the mean reciprocal rank),
and (iii) \texttt{Recall@\{1,2\}}.

\paragraph{Experimental Procedure}
We used the following procedure:
(i) train the model with a set of context-response pairs $\mathcal D = \{(x_i,y_i)\}_{i=1}^n$;
(ii) for each context sentence $x$ in the test data, rank the candidate responses $\{y_j\}_{j=1}^{10}$ by the model; and
(iii) report three evaluation metrics.

\paragraph{Baseline Measures}
As baseline measures, both (1) an RNN language model $\widehat{\mathbf P}_{\mathrm{RNN}}(y)$ \cite{Mikolov2010} and (2) a conditional RNN language model $\widehat{\mathbf P}_{\mathrm{RNN}}(y|x)$ \cite{Sutskever2014} were trained, and (3) PMI based on these language models, \textbf{RNN-PMI}, was also used for experiments~(see Equation~\eqref{eq:pmi_rnn}).
We trained these models with all combinations of the following settings:
(a) the number of dimensions of the hidden layers being $300$ or $1200$
and
(b) the initialization of the embedding layer being \textbf{\texttt{random}} (uniform on $[-0.1,0.1]$) or \textbf{\texttt{fastText}}.
For more detailed settings,
see Appendix \ref{sec:settings_rnn}.

\subsubsection*{Experimental Results}
\paragraph{Learning Time}
\label{sec:results_learning_time}
Table \ref{table:learning_time} shows the experimental results of the learning time\footnote{%
The computing environment was as follows:\newline
(i) CPU: Xeon E5-1650-v3 (3.5 GHz, 6 Cores);\newline
(ii) GPU: GTX 1080 (8 GB).%
}.
\begin{table}[tb]
\begin{center}
\tabcolsep 1pt %
\footnotesize
\begin{tabular}{l cc l rrrr}
\toprule
& \multicolumn{2}{c}{\multirow{2}{*}[-2pt]{\textit{Config}}}
& & \multicolumn{4}{c}{\textbf{Size of Training Set} $n$}
\\[2pt]
& & &
& \multicolumn{1}{r}{$10^3$\,}
& \multicolumn{1}{r}{$10^4$\,}
& \multicolumn{1}{r}{$10^5$\,}
& \multicolumn{1}{r}{$5\!\times\! 10^5$}
\\
\midrule
& \hspace{-3pt}\textit{Dim.} & \hspace{-5pt}\textit{Init.} \\[2pt]
\multirow{6}{*}{\rotatebox[origin=c]{90}{\textbf{RNN-PMI}\hspace{12px}}\hspace{4pt}}
& \multirow{3}{*}[-2pt]{\hspace{-3pt}300} & \multirow{3}{*}[-2pt]{\hspace{-6pt}\texttt{fastText}}
& Total
& 20.6
& 99.2
& 634.3
& 4042.5
\\ 
& & & \hspace{2pt} $\widehat{\mathbf P}(y)$
& 8.0
& 23.6
& 294.6
& 1710.1
\\
& & & \hspace{2pt} $\widehat{\mathbf P}(y|x)$
& 12.6
& 75.6
& 339.7
& 2332.4
\\ 
\cmidrule{2-8}
& \multirow{3}{*}[-2pt]{\hspace{-3pt}1200} & \multirow{3}{*}[-2pt]{\hspace{-6pt}\texttt{fastText}}
& Total
& 49.0
& 162.0
& 1751.3
& 13054.9
\\
& & & \hspace{2pt} $\widehat{\mathbf P}(y)$
& 16.3
& 57.2
& 671.0
& 5512.1
\\
& & & \hspace{2pt} $\widehat{\mathbf P}(y|x)$
& 32.7
& 104.8
& 1080.3
& 7542.8
\\
\midrule
\multirow{3}{*}{\rotatebox[origin=c]{90}{\textbf{PHSIC}\hspace{8px}}\hspace{4pt}} & \hspace{6pt}\textit{Encoder} & \textit{Kernel}\hspace{-2pt} \\[2pt]
& \multirow{1}{*}{\hspace{-2pt}\texttt{fastText}\hspace{-8pt}} & \hspace{2pt}\texttt{cos} & %
& \textbf{0.0}
& \textbf{0.1}
& \textbf{0.5}
& \textbf{2.8}
\\
\cmidrule{2-8}
& \multirow{1}{*}{\hspace{6pt}\texttt{DAN}} & \hspace{2pt}\texttt{cos} & %
& \textbf{0.0}
& \textbf{0.1}
& \textbf{0.4}
& \textbf{1.8}
\\
\bottomrule
\end{tabular}
\end{center}
\caption{\textbf{Learning time} [s] for each model and each size of training set for the dialogue response task.
Each row denotes a model; each column denotes the number of training data $n$.
The text appended to each baseline model denotes the number of dimension of hidden layers (\textit{Dim.}) and the method of initialization the embedding layer (\textit{Init.}).
The text appended to each proposed model denotes the pre-trained models used to encode sentences into vectors (\textit{Encoder}) and the kernel between these vectors (\textit{Kernel}).
The best result (the shortest learning time) in each column
is in bold.
}
\label{table:learning_time}
\end{table}%
Regardless of the size of the training set $n$,
the learning time for PHSIC is much shorter than that of the RNN-based method.
For example, even when the size of the training set $n$ is $5\times 10^5$, PHSIC is approximately $1400$--$4000$ times faster than RNN-based PMI.
This is because the estimators of PHSIC are reduced to a deterministic and efficient matrix calculation (Section \ref{sec:estimator_of_phsic}), whereas neural network-based models involve the sequential optimization of parameters via gradient descent methods.

\begin{table*}[tb]
\begin{center}
\setlength{\tabcolsep}{6pt} %
\footnotesize
\begin{tabular}{l cc cccc}
\toprule
\multirow{2}{*}[-2pt]{\textbf{Models}} & \multicolumn{2}{c}{\multirow{2}{*}[-2pt]{\textit{Config}}}
& \multicolumn{4}{c}{\textbf{Size of Training Set} $n$} \\[2pt]
& &
& \multicolumn{1}{c}{$10^3$}
& \multicolumn{1}{c}{$10^4$}
& \multicolumn{1}{c}{$10^5$}
& \multicolumn{1}{c}{$5\times 10^5$}
\\
\midrule
\multicolumn{3}{l}{Chance Level}
& .50; .29; .10, .20 %
& .50; .29; .10, .20 %
& .50; .29; .10, .20 %
& .50; .29; .10, .20 %
\\
\midrule
& \textit{Dim.} & \textit{Init.}
\\[3pt]
$\widehat{\mathbf P}_{\mathrm{RNN}}^{}(y)$ & 1200 & \texttt{fastText}
& .50; .29; .10, .21 %
& .50; .30; .11, .21 %
& .50; .30; .10, .21 %
& .50; .30; .13, .25 %
\\
\cmidrule(r){1-3} \cmidrule{4-7}
$\widehat{\mathbf P}_{\mathrm{RNN}}^{}(y|x)$ & 1200 & \texttt{fastText} 
& .50; .29; .10, .21 %
& .50; .30; .10, .21 %
& .52; .31; .11, .23 %
& .54; .32; .13, .25 %
\\
\cmidrule(r){1-3} \cmidrule{4-7}
\multirow{4}{*}[-2pt]{\textbf{RNN-PMI}}
& \multirow{2}{*}{300} & \texttt{random}
& .51; .30; .10, .21 %
& .51; .30; .11, .22 %
& .58; .35; .14, .29 %
& .69; .46; .25, .42 %
\\
& & \texttt{fastText} 
& .51; .29; .09, .20 %
& .56; .34; .15, .25 %
& .66; .41; .20, .36 %
& .76; .56; .36, .54 %
\\
\cmidrule(r){2-3} \cmidrule{4-7}
& \multirow{2}{*}{1200} & \texttt{random}
& .50; .29; .11, .20 %
& .51; .30; .10, .19 %
& .57; .35; .14, .29 %
& .70; .47; .26, .44 %
\\
& & \texttt{fastText} 
& .51; .30; .11, .20 %
& .52; .32; .13, .23 %
& .65; .42; .21, .36 %
& .75; .54; .34, .52 %
\\
\midrule
& \textit{Encoder} & \textit{Kernel}
\\[3pt]
\multirow{2}{*}[-2pt]{\textbf{PHSIC}}
& \multirow{1}{*}{\texttt{fastText}} & \texttt{cos}
& .61; .38; .17, .33 %
& .62; .40; .19, .34 %
& .62; .40; .19, .34 %
& .62; .40; .19, .34 %
\\
\cmidrule(r){2-3} \cmidrule{4-7}
& \multirow{1}{*}{\texttt{DAN}} & \texttt{cos}
& \textbf{.77}; \textbf{.58}; \textbf{.40}, \textbf{.56} %
& \textbf{.78}; \textbf{.57}; \textbf{.39}, \textbf{.56} %
& \textbf{.78}; \textbf{.58}; \textbf{.41}, \textbf{.57} %
& \textbf{.78}; \textbf{.58}; \textbf{.40}, \textbf{.57} %
\\
\bottomrule
\end{tabular}
\end{center}

\caption{\textbf{Predictive performance} for each model and each training set size for the dialogue response selection task:
\texttt{ROC-AUC}; \texttt{MRR}; \texttt{Recall@1,2}.
The best result 
in each column is in bold.
The other notation is the same as in Table \ref{table:learning_time}.
}
\label{table:experimental_results_predictive_performance}
\end{table*}
\paragraph{Robustness to Data Sparsity}
Table \ref{table:experimental_results_predictive_performance} shows the experimental results of the predictive performance.
When the size of the training data is small ($n\!=\!10^3, 10^4$),
that is, when the data is extremely sparse,
the predictive performance of PHSIC hardly deteriorates while that of PMI rapidly decays as the number of data decreases.
This indicates that PHSIC is more robust to data sparsity than RNN-based PMI owing to the effect of kernels.
Moreover,
PHSIC with the simple cosine kernel outperforms the RNN-based model regardless of the number of data,
while the learning time of PHSIC is thousands of times shorter than those of the baseline methods (Section \ref{sec:results_learning_time}).
Additionally we report Spearman's rank correlation coefficient between models to verify whether PHSIC shows similar behavior to PMI. See Appendix \ref{sec:correlation_between_models} for more detail.

\subsection{Data Selection for Machine Translation}
\label{sec:experiment_data_selection}
The aim of our second experiment was to demonstrate that PHSIC is also beneficial as a criterion of data selection.
To achieve this, we attempted to apply PHSIC to a parallel corpus filtering task that has been intensively discussed in recent (neural) machine translation (MT, NMT) studies.
This task was first adopted as a shared task in the third conference on machine translation (WMT 2018)\footnote{\url{http://www.statmt.org/wmt18/parallel-corpus-filtering.html}}.

Several existing parallel corpora, especially those automatically gathered from large-scale text data, such as the Web, contain unacceptable amounts of noisy (low-quality) sentence pairs that greatly affect the translation quality. %
Therefore, the development of an effective method for parallel corpus filtering would potentially have a large influence on the MT community;
discarding such noisy pairs may improve the translation quality and shorten the training time.

We expect PHSIC to give low scores to \emph{exceptional} sentence pairs (misalignments or missing translations) during the selection process because PHSIC assigns low scores to pairs that are highly inconsistent with other pairs (see Section~\ref{sec:phsic_as_smoothed_pmi}).
Note that applying RNN-based PMI to a parallel corpus selection task is unprofitable since obtaining RNN-based PMI also has an identical computational cost for training a sequence-to-sequence model for MT, and thus, we cannot expect a reduction of the total training time.

\subsubsection*{Experimental Settings}
\paragraph{Dataset}
We used the ASPEC-JE corpus\footnote{\url{http://lotus.kuee.kyoto-u.ac.jp/ASPEC/}}, which is an official dataset used for the MT-evaluation shared task held in the fourth workshop on Asian translation (WAT 2017)\footnote{\url{http://lotus.kuee.kyoto-u.ac.jp/WAT/WAT2017/}}~\cite{nakazawa-EtAl:2017:WAT2017}.
ASPEC-JE consists of approximately three million (3M) Japanese--English parallel sentences from scientific paper abstracts.
As discussed by~\citet{kocmi-varivs-bojar:2017:WAT2017}, ASPEC-JE contains many low-quality parallel sentences that have the potential to significantly degrade the MT quality.
In fact, they empirically revealed that using only the reliable part of the training parallel corpus significantly improved the translation quality.
Therefore, ASPEC-JE is a suitable dataset for evaluating the data selection ability.

\paragraph{Model}
For our data selection evaluation, we selected the Transformer architecture~\cite{Vaswani2017} as our baseline NMT model, which is widely-used in the NMT community and known as one of the current state-of-the-art architectures.
We utilized \texttt{fairseq}\footnote{\url{https://github.com/pytorch/fairseq}}, a publicly available tool for neural sequence-to-sequence models, for building our models.

\paragraph{Experimental Procedure}
We used the following procedure for this evaluation: 
(1) rank all parallel sentences in a given parallel corpus according to each criterion,
(2) extract the top $K$ ranked parallel sentences,
(3) train the NMT model using the extracted parallel sentences, and
(4) evaluate the translation quality of the test data using a typical MT automatic evaluation measure, i.e., BLEU~\cite{papineni-EtAl:2002:ACL}\footnote{We used {\tt multi-bleu.perl} in the Moses tool (\url{https://github.com/moses-smt/mosesdecoder}).
}.
In our experiments we evaluated PHSIC with $K=$ 0.5M and 1M.

\paragraph{Baseline Measure}
As a baseline measure, we utilize a publicly available script\footnote{\url{https://github.com/clab/fast_align}} of \textbf{fast\_align} \cite{Dyer2013}, which is one of the state-of-the-art word aligner. %
We firstly used the fast\_align for the training set $\mathcal D = \{(x_i,y_i)\}_i$ to obtain the word alignment between each sentence pair $(x_i,y_i)$, i.e., a set of aligned word pairs with its probabilities. We then computed the co-occurrence score of $(x_i,y_i)$ with sentence-length normalization, i.e., the average log probability of aligned word pairs.

\begin{table}[tb]
\begin{center}
\footnotesize
\tabcolsep 3.pt
\begin{tabular}{lccc ccc}
\toprule
\multicolumn{3}{l}{\multirow{2}{*}[-2pt]{\textbf{Selection Criteria}}} & \multicolumn{3}{c}{\hspace{-4pt}\textbf{\# of Selected Data} $K$} \\[2pt]
& & & 0.5M & 1M & 3M \\
\midrule
\multicolumn{3}{l}{(all the training set)\hspace{-5pt}} & - & - & 41.02 \\ %
\cmidrule{1-6}
\multicolumn{3}{l}{Random} & 34.26 & 39.82 & - \\ %
\cmidrule{1-6}
\multicolumn{3}{l}{\textbf{fast\_align}} & 38.63 & 40.56 & - \\
\cmidrule{1-6}
& \textit{Encoder} & \textit{Kernel}
\\[2pt]
\textbf{PHSIC}{\hspace{4pt}}
& \texttt{fastText} & \texttt{RBF} &\bf 38.95 &\bf 40.95 & - \\
\bottomrule
\end{tabular}
\end{center}
\caption{\textbf{BLEU scores}
with the Transformer
for each data selection criterion and each size of selected data $K$ for the parallel corpus filtering task.%
``Random'' represents the baseline method of selecting sentences at random.
}
\label{table:result_NMT}
\end{table}

\subsubsection*{Experimental Results}

Table~\ref{table:result_NMT} shows the results of our data selection evaluation. %
It is common knowledge in NMT that more data gives better performance in general.
However, we observed that PHSIC successfully extracted beneficial parallel sentences from the noisy parallel corpus; the result using 1M data extracted from the 3M corpus by PHSIC was almost the same as that using 3M data (the decrease in the BLEU score was only $0.07$),
whereas that by random extraction reduced the BLEU score by $1.20$.

This was actually a surprising result because PHSIC utilizes only monolingual similarity measures (kernels) without any other language resources.
This indicates that PHSIC can be applied to a language pair poor in parallel resources.
In addition, the surface form and grammatical characteristics between English and Japanese are extremely different%
\footnote{%
For example, word order; English is an SVO (subject-verb-object) language and Japanese is an SOV (subject-object-verb) language.%
}%
;
therefore, we expect that PHSIC will work well regardless of the similarity of the language pair.

\section{Related Work}
\paragraph{Dependence Measures}
Measuring independence or dependence (correlation) between two random variables, i.e., estimating dependence from a set of paired data, is a fundamental task in statistics and a very wide area of data science.
To measure the complex nonlinear dependence that real data has, we have several choices.

First, information-theoretic MI \cite{Cover2006} and its variants \cite{Suzuki2009,Rashef2011} are the most commonly used dependence measures.
However, to the best of our knowledge,
there is no practical method of computing MIs for large-multi class high-dimensional (having a complex generative model) discrete data, such as sparse linguistic data.

Second, several kernel-based dependence measures have been proposed for measuring nonlinear dependence \cite{Akaho2001,Bach2002,Gretton2005}.
The reason why kernel-based dependence measures work well for real data is that they do not explicitly estimate densities, which is difficult for high-dimensional data.
Among them, HSIC \cite{Gretton2005} is popular because it has a simple estimation method, which is used for various tasks such as feature selection~\citep{Song2012}, dimensionality reduction~\citep{Fukumizu2009}, and unsupervised object matching~\citep{Quadrianto2009,Jagarlamudi2010}.
We follow this line.

\paragraph{Co-occurrence Measures}
First, In NLP, PMI \cite{Church1989} and its variants \cite{Bouma2009} are the de facto co-occurrence measures between \emph{dense} linguistic expressions, such as words \cite{Bouma2009}
and simple narrative-event expressions \cite{Chambers2008}.
In recent years, positive PMI (PPMI) has played an important role as a component of word vectors \cite{Levy2014}.

Second, there are several studies
in which the pairwise ranking problem has been solved
by using deep neural networks (DNNs) in NLP.
\citet{Li2016} proposed a PMI estimation using RNN language models;
this was used as a baseline model in our experiments~(see Section \ref{sec:experiment_dialogue}).
Several studies have used DNN-based binary classifiers
modeling $\mathbf P(C=\text{positive}\mid(x,y))$ to solve the given ranking problem directly \cite{Hu2014,Yin2016,Mueller2016} (these networks are sometimes called Siamese neural networks).
Our study focuses on comparing co-occurrence measures.
It is unknown whether Siamese NNs capture the co-occurrence strength;
therefore we did not deal with Siamese NNs in this paper.

Finally, to the best of our knowledge,
\citet{Yokoi2017}'s paper is the first study that suggested converting HSIC to a pointwise measure.
The present study was inspired by their suggestion; here, we have (i) provided a formal definition (population) of PHSIC; (ii) analyzed the relationship between PHSIC and PMI; (iii) proposed linear-time estimation methods; and (iv) experimentally verified the computation speed and robustness to data sparsity of PHSIC for practical applications.

\section{Conclusion}
\label{sec:conclusion}
The NLP community has commonly employed PMI to estimate the co-occurrence strength between linguistic expressions;
however, existing PMI estimators have
a high computational cost
when applied to sparse linguistic expressions
(Section \ref{sec:introduction}).
We proposed a new kernel-based co-occurrence measure, the pointwise Hilbert--Schmidt independent criterion (PHSIC).
As well as defining PMI as the contribution to mutual information, PHSIC is defined as the contribution to HSIC; PHSIC is intuitively a ``kernelized variant of PMI'' (Section \ref{sec:definition_phsic}).
PHSIC can be applied to sparse linguistic expressions owing to the mechanism of smoothing by kernels.
Comparing the estimators of PMI and PHSIC,
PHSIC can be interpreted as a smoothed variant of PMI,
which allows various similarity metrics to be plugged in as kernels (Section \ref{sec:phsic_as_smoothed_pmi}).
In addition, PHSIC can be estimated in linear time owing to the efficient matrix calculation,
regardless of whether we use linear or nonlinear kernels (Section \ref{sec:estimator_of_phsic}).
We conducted a ranking task for dialogue systems and a data selection task for machine translation (Section~\ref{sec:experiments}).
The experimental results show that (i) the learning of PHSIC was completed thousands of times faster than that of the RNN-based PMI while outperforming it in ranking accuracy (Section~\ref{sec:experiment_dialogue});
and (ii) even when using a nonlinear kernel, PHSIC can be applied to a large dataset.
Moreover, PHSIC reduces the amount of training data to one third without sacrificing the output translation quality
(Section \ref{sec:experiment_data_selection}).

\paragraph{Future Work}
Using the PHSIC estimator in feature space (Equation \eqref{eq:phsic_RKHS}), we can generate the most appropriate $\psi(y)$ for a given $\phi(x)$ (uniquely, up to scale).
That is,
if a DNN-based sentence decoder is used, $y$ (a sentence) can be restored from $\psi(y)$ (a feature vector) so that generative models of strong co-occurring sentences can be realized.

\section*{Acknowledgments}
We are grateful to anonymous reviewers for their helpful comments.
We also thank
Weihua Hu for useful discussions,
Kenshi Yamaguchi for collecting data, and
Paul Reisert for proofreading.
This work was supported in part by JSPS KAKENHI Grant Number JP15H01702 and JST CREST Grant Number JPMJCR1513, Japan.

\clearpage
\bibliography{Mendeley,nmt}
\bibliographystyle{acl_natbib_nourl}

\clearpage
\appendix

\section{Available Kernels for PHSIC}
\label{sec:available_kernels_appendix}
\paragraph{Similarity between Sentence Vectors}
A variety of vector representations of phrases and sentences based on the distributional hypothesis have recently been proposed,
including sentence encoders~\cite{Kiros2015,Dai2015,Iyyer2015,Hill2016,Cer2018} and the sum of word embeddings;
it is known as \emph{additive compositionality}~\cite{MitchellLapata2010,Mikolov2013,Wieting2015} that we can express the meaning of phrases and sentences well with the sum of word vectors (e.g., \texttt{word2vec}~\cite{Mikolov2013a}, \texttt{GloVe}~\cite{Pennington2014a}, and \texttt{fastText}~\cite{Bojanowski2017}).
Note that various pre-trained models of sentence encoders and word embeddings have also been made available.

The cosine of these vectors, which is a positive definite kernel, can be used as a convenient and highly accurate similarity function between phrases or sentences.
Other major kernels can also be used, such as the RBF kernel, the Laplacian kernel, and polynomial kernels.

\paragraph{Structured Kernels}
Various structured kernels for NLP,
such as tree kernels,
which capture fine structure of sentences such as syntax,
were devised in the support vector machine era~\cite{Collins2002,Bunescu2006,Moschitti2006}.

\paragraph{Combinations}
We can freely combine the previously mentioned kernels
because the sum and the product of positive definite kernels are also positive definite kernels~\cite[Proposition 3.22]{Shawe-TaylorCristianini2004}.

\section{Derivation of Fast PHSIC Estimation in Data Space}
\label{sec:estimation_data_space_appendix}

Although estimators of HSIC and PHSIC depend on kernels $k, \ell$ and data $\mathcal D$,
hereinafter, we use the following notation for the sake of simplicity:
\begin{align}
& \widehat{\mathrm{HSIC}}(X,Y) := \widehat{\mathrm{HSIC}}(X,Y;\mathcal D,k,\ell)
\text{,}
\\
& \widehat{\mathrm{PHSIC}}(x,y) := \widehat{\mathrm{PHSIC}}(x,y;\mathcal D,k,\ell)
\text{.}
\end{align}

\paragraph{Na\"{i}ve Estimation}
Fist, an estimator of PHSIC in the data space~\eqref{eq:phsic_population_kernel} is
\begin{align}
\!\widehat{\mathrm{PHSIC}}_{\mathrm{kernel}}(x,y)
\!=\! \textstyle (\VEC{k} - \overline{\VEC{k}})^{\!\top\!} (\frac 1 n H) (\VEC{\ell} - \overline{\VEC{\ell}})\label{eq:phsic_kernel}
\text{,}\!\!
\end{align}
where $\VEC{k} := (k(x,x_1),\dots,k(x,x_n))^\top\in\mathbb R^n$, so as $\VEC{\ell}$;
and vector $\overline{\VEC{k}} := \frac 1 n K \VEC{1}$ denotes empirical mean of $\{\VEC{k}_i\}_{i=1}^n$,
so as $\overline{\VEC{\ell}}$.
This estimation has a large computational cost.
When learning,
computing the vectors $\overline{\VEC{k}}, \overline{\VEC{\ell}}$ takes $\mathcal O(n^2)$ time and $\mathcal O(n)$ space.
When estimating PHSIC,
computing $\VEC{k},\VEC{\ell}$ and multiplying the matrix $\frac 1 n H$ takes $\mathcal O(n)$ time.

\paragraph{Fast Estimation via Incomplete Cholesky Decomposition}
Equation~\eqref{eq:phsic_kernel} has a large computational cost because
it is necessary to construct the Gram matrices $K$ and $L\in\mathbb R^{n\times n}$.
In kernel methods, several methods have been proposed for approximating Gram matrices at low cost without constructing them explicitly, such as \emph{incomplete Cholesky decomposition}~\cite{Fine2001}.

By incomplete Cholesky decomposition,
from data points $\{x_1,\dots,x_n\}\subseteq \mathcal X$ and a positive definite kernel $k\colon\mathcal X\times\mathcal X\to\mathbb R$,
a matrix $A = (\VEC{a}_1,\dots,\VEC{a}_n)^\top\in\mathbb R^{n\times d}$~($d\ll n$) can be obtained with $\mathcal O(nd^2)$ time complexity.
This makes it possible to approximate the Gram matrix $K$ by vectors $\VEC{a}_i\in\mathbb R^d$ without configuring the entire of $K$:
\begin{align}
\VEC{a}_i^\top\VEC{a}_j &\approx k(x_i,x_j)\\
AA^\top &\approx K\text{.}
\end{align}

Also, for HSIC, an efficient approximation method utilizing incomplete Cholesky decomposition has been proposed~\cite[Lemma~2]{Gretton2005}:
\begin{align}
\widehat{\mathrm{HSIC}}_{\mathrm{ICD}}(X,Y)
&= \frac 1 {n^2} \lVert(HA)^\top B\rVert_{\mathrm{F}}^2\label{eq:hsic_icd}
\text{,}
\end{align}
where $A = (\VEC{a}_1,\dots,\VEC{a}_n)^\top\in\mathbb R^{n\times d}$ is a matrix satisfying $AA^\top\approx K$ computed via incomplete Cholesky decomposition, so as $B$~($BB^\top\approx L$).
Equation~\eqref{eq:hsic_icd} can be represented in the form of the expectation on data points:
\begin{align}
& \widehat{\mathrm{HSIC}}_{\mathrm{ICD}}(X,Y)
\!=\!
\frac 1 n \!\sum_{i=1}^n
\Bigl[
\empheqpos{}{\!
\textstyle
(\VEC{a}_i \!-\! \overline{\VEC{a}})^{\!\top}
\widehat C_{\mathrm{ICD}}^{}
(\VEC{b}_i \!-\! \overline{\VEC{b}})
\!}
\Bigr]
\label{eq:hsic_icd_bilinear}\\[-8pt]
&
\widehat C_{\mathrm{ICD}}^{} := \frac 1 n (HA)^\top B \in \mathbb R^{d\times d}
\text{,}
\end{align}
where vector $\overline{\VEC{a}} := \frac 1 n A^\top\VEC{1} \in\mathbb R^d$ denotes empirical mean of $\{\VEC{a}_i\}_{i=1}^n$, so as $\overline{\VEC{b}} := \frac 1 n B^\top\VEC{1}$.

Recall that $\mathrm{PHSIC}(x,y)$ is the contribution of $(x,y)$ to $\mathrm{HSIC}(X,Y)$~(see Section~\ref{sec:hsic_and_phsic});
PHSIC then can be efficiently estimated by the shaded part of Equation~\eqref{eq:hsic_icd_bilinear}:
\begin{align}
\textstyle
\widehat{\mathrm{PHSIC}}_{\mathrm{ICD}}(x,y)
\!=\!
\empheqpos{}{\!
(\VEC{a} \!-\! \overline{\VEC{a}})^\top
\widehat C_{\mathrm{ICD}}^{}
(\VEC{b} \!-\! \overline{\VEC{b}})
\!}\label{eq:phsic_icd}
\,\text{.}\!
\end{align}
Here,
the vector $\VEC{a}\in\mathbb R^d$ corresponding to the new $x$ can be calculated by ``performing from halfway'' on the incomplete Cholesky decomposition algorithm.
Let $x^{(1)},\dots,x^{(d)}$ denote the dominant $x_i$s adopted during decomposition algorithm.
The $j$th element of $\VEC{a}$ can be computed as follows:
\begin{align}
\VEC{a}[j] \!=\! \Bigl[
k(x,x^{(j)}) - \sum_{m=1}^{j-1}\VEC{a}[m] A_{jm}
\Bigr] \,/\, A_{jj}
\text{,}
\end{align}
so as $\VEC{b} \in \mathbb R^d$ corresponding to the new $y$.
The estimation via incomplete Cholesky decomposition~\eqref{eq:phsic_icd}
is extremely efficient compared to the naive estimation~\eqref{eq:phsic_kernel};
Equation~\eqref{eq:phsic_icd}'s computational complexity is equivalent to the estimation in the feature space~\eqref{eq:phsic_RKHS}.

\section{Detailed Settings for Learning RNNs}
\label{sec:settings_rnn}
Detailed settings for learning RNNs used in this research are as follows.
\begin{itemize}
\item Hidden layers: single layer LSTMs \cite{Hochreiter1997}
\item Vocabulary: words with a frequency: $10$ or more ($n=5\times 10^5$), $2$ or more (otherwise)
\item Dropout rate: $0.1$ ($300$-dim), $0.3$ ($1200$-dim)
\item Batch size: $64$
\item Max epoch number: $5$ ($n=5\times 10^5$), $30$ (otherwise)
\item Deep learning framework: \texttt{Chainer} \cite{Tokui2015}
\end{itemize}

\section{Correlation Between Models\newline in Dialogue Response Selection Task}
\label{sec:correlation_between_models}
Table \ref{table:correlation_between_models} shows Spearman's rank correlation coefficient (Spearman's $\rho$) between the co-occurrence scores on the test set computed by the models in the dialogue response selection task (Section \ref{sec:experiment_dialogue}).
\begin{table}[tb]
\label{table:correlation_between_models}
\begin{center}
\setlength{\tabcolsep}{2pt} %
\footnotesize
\begin{tabular}{l ccc cccc}
\toprule
\multirow{2}{*}[-2pt]{\textbf{Models}} & \multicolumn{2}{c}{\multirow{2}{*}[-2pt]{\textit{Config}}}
& & \multicolumn{4}{c}{} \\[2pt]
& & &
& \multicolumn{1}{c}{\texttt{(A)}}
& \multicolumn{1}{c}{\texttt{(B)}}
& \multicolumn{1}{c}{\texttt{(C)}}
& \multicolumn{1}{c}{\texttt{(D)}}
\\
\midrule
& \textit{Dim.} & \textit{Init.}
\\[3pt]
\multirow{2}{*}[-2pt]{\textbf{RNN-PMI}}
& \multirow{1}{*}{300} & \texttt{fastText} & \hspace{0.5em}\texttt{(A)}\hspace{0.5em}
& --
& .42
& .12
& .27
\\
\cmidrule(r){2-4} \cmidrule{5-8}
& \multirow{1}{*}{1200} & \texttt{fastText} & \hspace{0.5em}\texttt{(B)}\hspace{0.5em}
& .42
& --
& .12
& .26
\\
\midrule
& \textit{Encoder} & \textit{Kernel}
\\[3pt]
\multirow{2}{*}[-2pt]{\textbf{PHSIC}}
& \multirow{1}{*}{\texttt{fastText}} & \texttt{cos} & \hspace{0.5em}\texttt{(C)}\hspace{0.5em}
& .12
& .12
& --
& .16
\\
\cmidrule(r){2-4} \cmidrule{5-8}
& \multirow{1}{*}{\texttt{DAN}} & \texttt{cos} & \hspace{0.5em}\texttt{(D)}\hspace{0.5em}
& .27
& .26
& .16
& --
\\
\bottomrule
\end{tabular}
\end{center}
\caption{\textbf{Spearman's $\rho$} between the co-occurrence scores computed by the models in the dialogue response selection task (Section \ref{sec:experiment_dialogue}).
The size of training set $n$ is $5\times 10^5$.
The other notation is the same as in Table \ref{table:learning_time}.}
\end{table}%
This shows that the behavior of RNN-based PMI and PHSIC are considerably different.
Furthermore, interestingly, the behavior of PHSICs using different kernels is also different.
Possible reasons for these observations are as follows:
(1) the difference in the dependence measures (MI or HSIC) on which each model is based;
(2) the validity or numerical stability of estimating PMI with RNN language models; and
(3) differences in the behavior of PHSIC originating from differences in the plugged in kernels.
A more detailed analysis of the compatibility between tasks and measures (or kernels) is attractive future work.

\end{document}